\documentclass[runningheads]{llncs}

\usepackage{amssymb}
\usepackage{amsfonts}
\usepackage{amsmath}
\usepackage{mathrsfs}
\usepackage{mathptmx}
\usepackage{mathrsfs}
\usepackage{bbm}
\usepackage{bm}
\usepackage{graphicx}
\usepackage{float}
\usepackage{booktabs}
\usepackage[linesnumbered,ruled]{algorithm2e}
\usepackage{cite}
\usepackage{array}
\usepackage{color}
\usepackage{url}
\usepackage{stfloats}
\usepackage[strings]{underscore}
\usepackage[marginal]{footmisc}

\newcommand{\tabincell}[2]{\begin{tabular}{@{}#1@{}}#2\end{tabular}}

\newcommand{\PreserveBackslash}[1]{\let\temp=\\#1\let\\=\temp}
\newcolumntype{C}[1]{>{\PreserveBackslash\centering}p{#1}}
\newcolumntype{R}[1]{>{\PreserveBackslash\raggedleft}p{#1}}
\newcolumntype{L}[1]{>{\PreserveBackslash\raggedright}p{#1}}

\makeatletter
\newcommand*{\centerfloat}{%
  \parindent \z@
  \leftskip \z@ \@plus 1fil \@minus \marginparwidth
  \rightskip \leftskip
  \parfillskip \z@skip}
\makeatother

\usepackage[colorlinks,linkcolor=black,anchorcolor=black,citecolor=black,urlcolor=blue]{hyperref}

\begin{document}

\setcounter{secnumdepth}{4}
\title{  A Novel Neural Network-based Symbolic Regression Method: Neuro-Encoded Expression Programming
}
\titlerunning{Neuro-Encoded Expression Programming}
%

\author{Aftab Anjum\inst{1} \and 
Fengyang Sun\inst{1} 
\and Lin Wang\inst{1}* 
\and Jeff Orchard \inst{2} 
}

\authorrunning{A. Anjum et al.}
%
\institute{
Shandong Provincial Key Laboratory of Network Based Intelligent Computing, University of Jinan, Jinan, 250022, China
\and David R. Cheriton School of Computer Science, University of Waterloo, Waterloo, Ontario N2L 3G1, Canada\\
*Corresponding Author: \email{wangplanet@gmail.com}
}
\maketitle              
\footnote{Aftab Anjum and Fengyang Sun ---- Contribute equally to this article. \\
\scriptsize This work has been published on \textit{2019 International Conference on Artificial Neural Networks}, pp. 373–386, 2019. The final authenticated publication is available online at \url{https://doi.org/10.1007/978-3-030-30484-3_31}}

\begin{abstract}
Neuro-encoded expression programming(NEEP) that aims to offer a novel continuous representation of combinatorial encoding for genetic programming methods is proposed in this paper.
Genetic programming with linear representation uses nature-inspired operators (e.g., crossover, mutation) to tune expressions and finally search out the best explicit function to simulate data.
The encoding mechanism is essential for genetic programmings to find a desirable solution efficiently.
However, the linear representation methods manipulate the expression tree in discrete solution space, where a small change of the input can cause a large change of the output.
The unsmooth landscapes destroy the local information and make difficulty in searching. The neuro-encoded expression programming constructs the gene string with recurrent neural network (RNN) and the weights of the network are optimized by powerful continuous evolutionary algorithms. The neural network mappings smoothen the sharp fitness landscape and provide rich neighborhood information to find the best expression. The experiments indicate that the novel approach improves training efficiency and reduces test errors on several well-known symbolic regression problems.
\keywords{Genetic Programming \and Symbolic Regression \and Neural Network \and Gene Expression Programming
 \and Evolutionary Algorithm.}
\end{abstract}

\section{Introduction}
\begin{figure}[htp]
    \centering
    \includegraphics[width=0.97\textwidth]{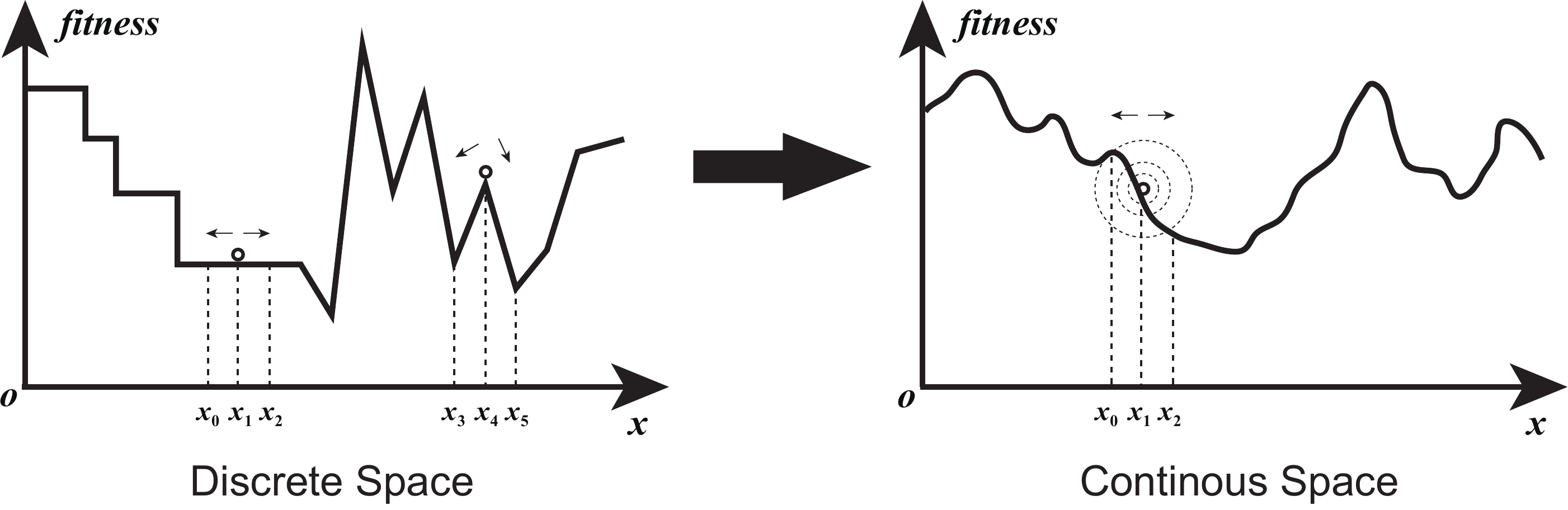}
    \caption{A sketch plot of comparison between discrete and continuous space. The discrete space shows two types of fitness landscape features. Plateau is that the fitness values around point $x_1$ is the same as central point. The neighborhood of $x_4$ is extremely sharp and fluctuant. Instead, the continuous space shows that the central point can obtain a slope in a small or large neighborhood.
    }
    \label{discrete-continous}
\end{figure}

Symbolic regression (SR) \cite{symbolicreg} is to find an explicit function for simulation of user-defined data. Compared to implicit numerical (linear or nonlinear) regression analysis, SR can construct a function for a complex data without any prior knowledge and generally has powerful interpretability due to clear mathematic formula.
Currently, genetic programming methods with linear representation \cite{Candida01,linergentic,mep} are mainly used to solve SR, which adopts a number of nature-inspired operators such as mutation and crossover to manipulate expressions and has presented decent performances on various applications.
However, the linear representation approaches encode the expressions in discrete manner, which considers it as a combinatorial problem. Compared to continuous problems, the combinatorial problem do not provide sufficient and useful neighborhood information to aid searching \cite{Maehara2018,Pardalos2006}. In addition, the local structure of the combinatorial problem are hard ``sharp'' style and the fitness landscape is not smooth where a minute change of the genotype can instill a substantial change of the phenotype \cite{Rothlauf2006,Fan2014Approximation} and may cause oscillation of converging process \cite{George2018Oscillation} (Fig. \ref{discrete-continous}). All the factors above make it difficult to find a desirable function fitting data for linear representation methods.

\par
Recently neural networks have achieved great success in generative tasks \cite{ganimage,ganmusic,wang2018textgen}. In particular, neural networks have demonstrated considerable potential for generating texts or strings \cite{bowman2016generating}. The genetic programming methods (such as gene expression programming) can decode a string to an expression tree that is equivalent to a mathematical function. Then the two facts make it possible to use neural networks to generate expressions, which converts the purely discrete encoding to continuous encoding and alleviates the aforementioned difficulties in solution space.

\par
Therefore, we propose a neuro-encoded expression programming (NEEP), which constructs the mathematical functions with neural networks generating expression string.
Instead of the discrete way, a small change in continuous weights vector only triggers similar form of function and makes slow-varying effects. Therefore, the continuous neural network mappings smoothen and soften the hard sharp discrete fitness landscape and provide more flexible local information.
In this manner, the NEEP can adopt powerful continuous optimizers to finely adjust the weights of network and find better function.

\section{Related Works}

\subsection{Symbolic Regression}

The purpose of symbolic regression is to find an explicit function, which is the primary difference with numerical regression. For a predefined data, SR finds an explicit function $f:{\bf{x}} \to y$ that approximates the data with minimum error.

\par
\subsection{Genetic Programming}
There are several types of methods that can be used to solve SR, e.g., analytic programming \cite{analyticprog}, fast function extraction \cite{ffx}, grammar evolution \cite{GE} and genetic programming \cite{geneticprog}. Genetic programming methods (such as gene expression programming and standard genetic programming) are one type of the commonly used methods to solve symbolic regression.
Standard genetic programming (GP) \cite{geneticprog} tunes the tree structure of expression directly by nature-inspired operators, e.g., crossover is the exchange in subtrees of two chromosomes at certain nodes. GPs maintain good patterns but suffer from the explosion of tree size.
Gene expression programming (GEP) \cite{Candida01,wang2015building} constructs chromosomes with linear expression strings and provides an efficient way to encode syntactically correct expressions. For readers' better understanding of our method, we introduce more details about GEP.

\par
Gene expression programming encodes the expression tree structure into fixed length linear chromosomes. Structurally, GEP genes consist of head and tail. Head consists of function symbols and terminal symbols and tail contains terminal symbols only. GEP uses the population of linear expression strings, selects them according to their fitness values and introduces genetic variation through genetic operators.

\par
The encoding design \cite{zhong2017gene} has a significant influence on the performance of gene expression programming since it determines the search space as well as the mapping between genotypes and phenotypes.
Traditional GEP adopts the K-expression representation \cite{Candida01}, which converts a linear string to an expression tree by using a breadth-first travelling procedure.
Li et al. \cite{Li01}introduced new enhancements (P-GEP), which improve the encoding design by suggesting the depth-first technique of converting the string into the expression Tree. P-GEP increases the searching efficiency, but it is not scalable for complex problems.
Automatically defined functions (ADF) \cite{kozaADF} were, for the first time, introduced by Koza as a way of reusing code in genetic programming. Ferreira \cite{Ferreira2006} introduced improvements (GEP-ADF) to encode the subfunctions into the expression tree which makes the GEP more flexible and robust.
However, these encoding improvements are still based on discrete space, which lacks of sufficient neighborhood information. To the best of our knowledge, there are few approaches encoding the expression string in a continuous manner. Therefore, the difficulties from discrete encoding are yet to be solved.

\subsection{Neural Network on Generation Task}

Neural network has achieved success on generation task such as image \cite{ganimage}, audio \cite{ganmusic} and text \cite{bowman2016generating} generation.
In particular, several studies on text/string generation by neural networks are reviewed in this part.
Bowman et al. \cite{bowman2016generating} proposed an RNN-based variational autoencoder (VAE) language model that incorporates distributed latent representations of entire sentences in a continuous space and explicitly models holistic properties of sentences.
Wang et al. \cite{wang2018textgen} generated texts based on generative adversarial networks (GANs). This method builds discriminator with a convolutional neural network and constructs generator with RNN and VAE to solve the problem that GANs always emit the similar data.

\begin{figure*}[htb]
    \centering
    \includegraphics[width=0.97\textwidth]{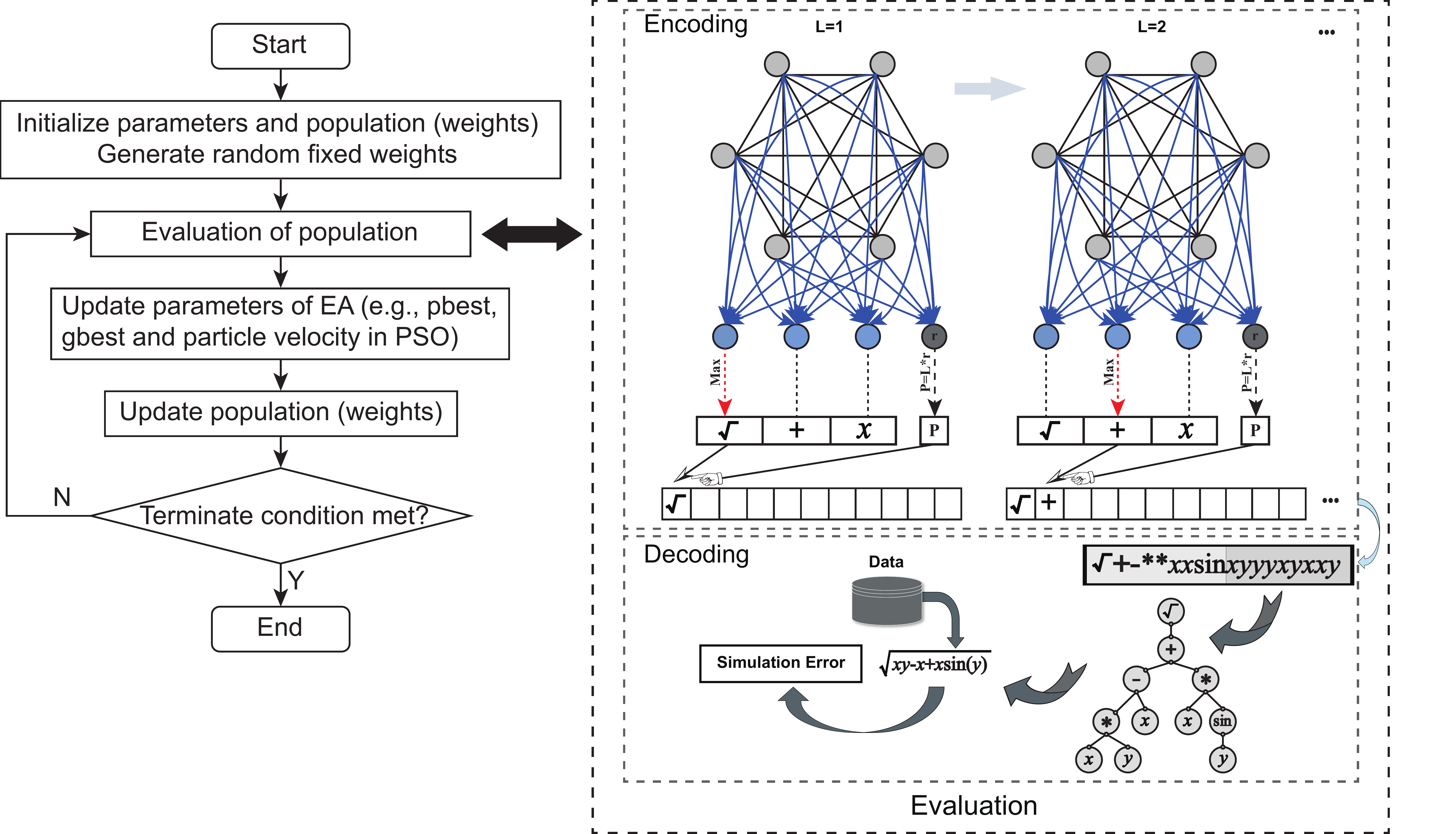}
    \caption{The framework for neuro-encoded expression programming in a general description. The black dashed square represents the fitness evaluation of the evolutionary algorithm. The neural encoding in evaluation is the main contribution of this study, which uses the neural network to encode the expression string. The network structure details is seen in Fig. \ref{neuralnetwork}. The first light-colored segment of linear string is the head part and the second part is tail. The string is decoded into a function in breadth-first scheme as GEP. The simulation error will return back to the evolutionary algorithm as the fitness value in optimization.}
    \label{overallstructure}
\end{figure*}

\par
To our knowledge, there is little research that uses neural network to solve symbolic regression by generating expression strings in spite of its success on generating texts. Liskowski et al. \cite{LiskowskiBK18} proposed a method in which neural networks play a ``pre-training'' role to detect possible patterns in data, and then aids in finding the function rather than generating expression strings directly.
Yin et al. \cite{YinGRN2012} proposed a novel self-organizing reservoir computing methods by gene regulatory network. This method can process arbitrary sequences of inputs such as speech recognition, whereas generating expression is not a sequential problem and there is no external ``formula'' input for training.
Another type of interesting works are word embedding methods \cite{levy2014neural}, however in reality we do not have enough formula data to learn their underlying similarity.

\section{Methodology}
The neuro-encoded expression programming (NEEP) adopts an evolutionary algorithm (EA) to optimize the network connection weights, then use the neural network to generate the expression string, and the K-expression method to decode the string into a function and then calculate the simulation error (Fig. \ref{overallstructure}). The main contribution of this paper is the method for encoding and generating the expression tree, which is based on the output of a neural network.

\subsection{General Framework}

\par
The first step of this method is initializing the parameters (e.g., pbest and gbest in particle swarm optimization \cite{pso}, distribution mean and step size in covariance matrix adaptation evolution strategy \cite{CMAES}) and population (all the net weights to be optimized) of the evolutionary algorithm.
When evaluating each network weights vector, the weights are inserted into a recurrent neural network that generates the expression strings composed of function operators (such as +, -, $*$, /) and terminal symbols (e.g., variables). After that, these strings are decoded into expression trees, which are equivalent to mathematical functions.
By putting the data into the expression, we can compare its value and to the target value. We use the resulting error as the fitness value of each individual network in population.
Then, we update all the necessary parameters in the evolutionary algorithm (e.g., update pbest, gbest and velocity according to fitness in PSO) and update the current solutions set (weights to be optimized). We repeat the above process until the termination condition is met (Fig. \ref{overallstructure}).

\subsection{Encoding}

\begin{figure}[htp]
    \centering
    \includegraphics[width=0.6\textwidth]{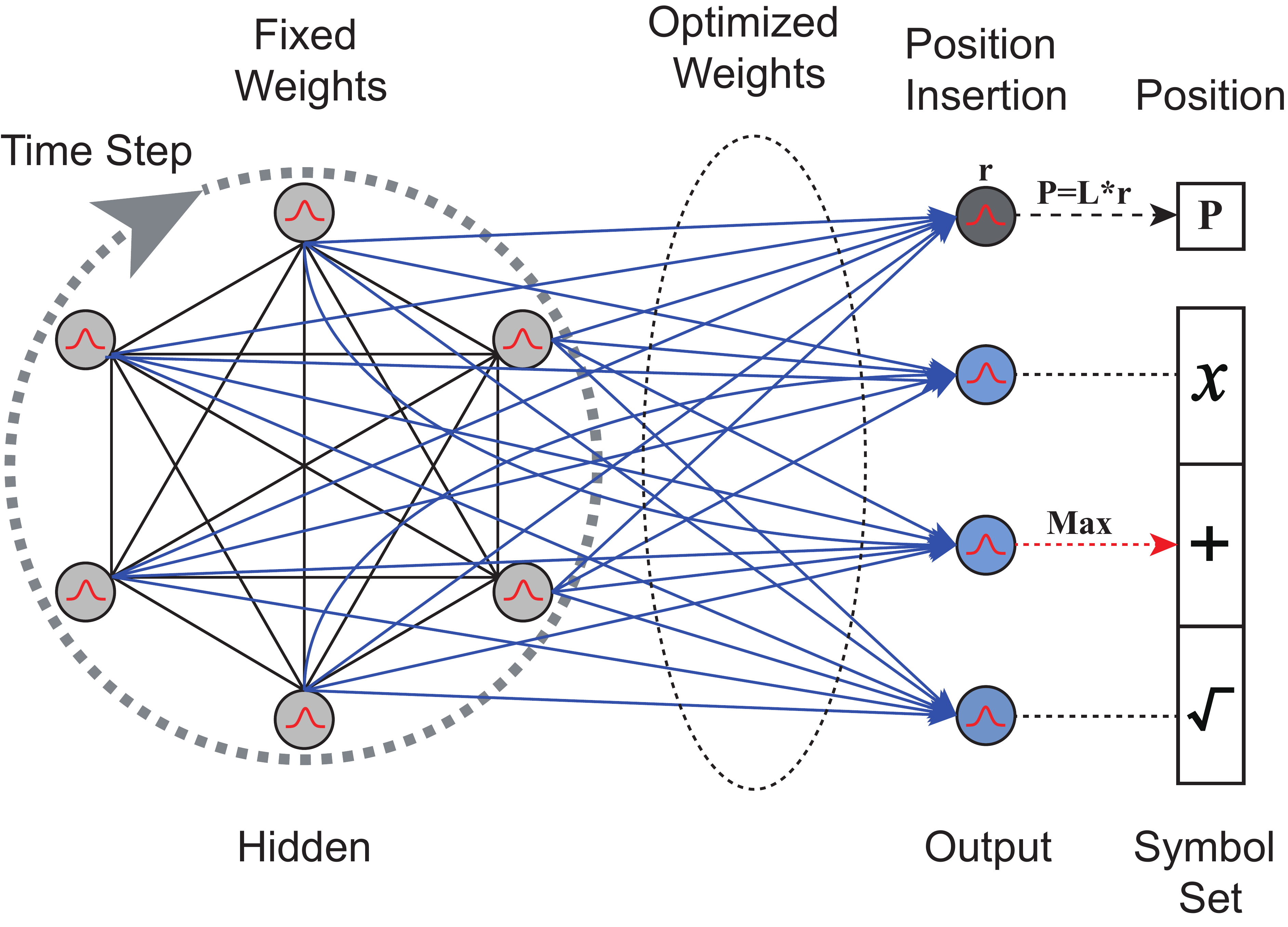}
    \caption{The architecture of the encoder demonstrated in Fig. \ref{overallstructure}. The black lines in the dashed circle represent the fixed hidden weights of the recurrent neural network. The blue arrows are the output weights to be updated. After all the time steps, the model obtains the outputs and each output neuron corresponds to a function or terminal symbol. Then the output neuron with maximum value will trigger the single symbol at certain position. $L$ is the current length of the expression string and $r$ is the output value as position rate in $[0,\ 1]$. $P$ denotes the position where the triggered symbol will be inserted. This general formula is interpreted into two cases of Eq. \ref{eq:posHead} and Eq. \ref{eq:posTail}.}
    \label{neuralnetwork}
\end{figure}

Inspired by biological brain, neural network presents flexible learning and powerful representation capability \cite{wang2017investigating} and has been widely designed and applied in various fields \cite{wang2017nnp,huang2018fwnn}.
The fully connected recurrent neural network  \cite{wang2017investigating} thus acts as the encoder which specifies the linear expression string.
The neural network consists of hidden neurons and the output neurons, and all the hidden neurons are fully connected with each other. As is shown in Fig. \ref{neuralnetwork}, there are no formal input neurons in the network because no external information is input into the network during the generation of the string. Instead of back propagation, we use an evolutionary algorithm to optimize the weights between the output neurons and all the hidden neurons (see optimization part). The Gaussian shape function $f=e^{-x^2}$ is chosen as the activation function of the hidden and output neurons instead of a sigmoid function because of premature convergence during encoding the string.
\par
Firstly, for symmetry, each neuron is initialized to zero.
The hidden weights are uniformly randomized before the evolution and keep constant during the evolution. This fixed weights are the same for all the neural networks.
As the behavior of the network can be chaotic, a small change in the initial condition could produce a significant difference in the later state \cite{manjunath2013echo}. As shown in \cite{ChaoticNN}, to reduce the instability we introduce fixed weights among all the hidden neurons and the remaining weights are still capable of finding the underlying pattern. All the other weights between the hidden and output neurons are uniformly randomized and are further optimized by evolutionary algorithms.
After each time step, the output neurons are updated.
Then, after all the time step the output neuron with the maximum value indicates which function or terminal symbol will be inserted into the expression string at a certain position.
The position is determined by an additional neuron in the output layer in default, naming it the position insertion neuron. The process keeps inserting the symbols into the expression string until the desired length is achieved.

\par
The string modeling is based on head and tail \cite{Candida01}. The number of output neurons is determined by the size of the function and terminal sets of specific problem. For position identification of head part symbols (terminals and functions), the position insertion in the head part of the string $p_\mathrm{h}$ can be assessed by Eq. \ref{eq:posHead}.
\begin{equation} \label{eq:posHead}
p_\mathrm{h} = \mathrm{round}(i_\mathrm{out} \cdot L+1 ),
\end{equation}
\noindent where $i_\mathrm{out}$ is output value of insertion neuron as position rate, which is distributed in (0, 1]. $L$ is the current length of the expression string. On the other hand, if $L$ is larger than the head size $h$ then, the corresponding terminal symbol will be inserted at a certain point of the tail part according to the value of position insertion neuron $i_\mathrm{out}$. The value of position insertion neuron in the tail part $p_\mathrm{t}$ can be calculated by the given Eq. \ref{eq:posTail}.
\begin{equation} \label{eq:posTail}
p_\mathrm{t} = \mathrm{round}(i_\mathrm{out} \cdot (L - h+1) + h ).
\end{equation}
The whole process of symbol injection can be seen the encoding part in Fig. \ref{overallstructure}.

\subsection{Decoding}
\par
The decoder is the translator which transfers the information from the string into the expression tree. The translation starting position is always the first position of the gene, whereas the last position of the gene does not necessarily coincide with the termination point.

\par
Let us consider the encoded gene ``$\sqrt{\ }\ $+-**\textit{xx}sin\textit{xyyyxyxxy}'' as represented in Fig. \ref{overallstructure}. This encoded gene can be translated into the expression tree by the breadth-first technique which is further decoded into the mathematical function.
The fitness value of each mathematical function/expression is calculated by measuring how well it fits the data, using mean square error (MSE) between the predicted values and the desired values. The decoding process is the same as in GEP (see more details in \cite{Candida01}).

\subsection{Optimization}

In the NEEP framework, we can choose different evolutionary algorithms to optimize the neural network for producing the most accurate expression. Three versions of NEEP are proposed in this work, which is GA-NEEP (based on GA, genetic algorithm \cite{holland1992adaptation}), PSO-NEEP (based on PSO, particle swarm optimization \cite{pso}), CMAES-NEEP (based on CMA-ES, covariance matrix adaptation evolution strategy \cite{CMAES}). In all the evolutionary algorithms, the population (chromosomes or particles) are the weight vectors, their values are uniformly randomized and then insert into the neural network for encoding the expression strings.
\par
We do not use back propagation (BP) because the problem is different from conventional supervised learning. The evolved function is not fixed during the calculation of derivative of weights. It is quite hard to obtain gradients with a uniform BP for all the problems and evolved functions. On the other hand, evolutionary algorithms \cite{chen2017iching} can conduct efficient optimization by approximating search direction without gradient information and become an appropriate choice of optimizer.

\subsubsection{Genetic Algorithm}

Genetic algorithm (GA) \cite{holland1992adaptation} is the search and optimization procedure inspired by the natural selection process of genetic. For evolving the population selection, crossover and mutation operators are used for creating better offspring than the parents.
GA is a robust optimization method that is easy to implement and requires few problem information.
In this study, the individuals are the weight vectors whose values are randomly created. Each weight vector is put into the corresponding neural network and then obtains its fitness value. GA is adopted in this work as a baseline method.
\subsubsection{Partials swarm optimization}

Partials swarm optimization (PSO) \cite{pso,wang2016tournamentpso} is a global numerical algorithm inspired by birds foraging. In PSO the particle migrate towards the direction of a combination of their former velocity personal best and global best and update its position. In this work, the particles are the weight vectors, and their values are randomly created. For the evaluation of these particles, they are inserted into the neural network and calculate their fitness values.
PSO has shown robustness and efficiency in finding global optima.

\subsubsection{Covariance matrix adaptation evolution strategy}

Covariance matrix adaptation evolution strategy (CMA-ES) \cite{CMAES} is a leading stochastic and derivative-free method for continuous optimization of non-linear, non-convex functions. The CMA-ES samples candidate points with a multivariate normal distribution. It updates the mean vector and covariance matrix so that it encourages reproducing previously successful search steps based on the maximum-likelihood principle. The CMA-ES has shown advantageous convergence property when compared to many other evolutionary algorithms for a wide class of problems. CMSA-ES is adopted in this study as one of the optimizers because of the mentioned advantages and powerful performances.

\section{Experiments}

\par
This section explores the performance of the proposed NEEP, and will not devise more sophisticated wrappers around GEP to improve the encoding way. The analysis will be limited to synthetic and benchmark regression problems. The three proposed methods are compared with standard GEP and standard GP. The problem configurations are outlined in Table 1, and the algorithm settings are described in the following subsection. Finally, the convergence behaviors and test accuracy are discussed.

\subsection{Benchmark Configurations}
We evaluated the proposed methods, GEP and GP on 14 synthetic benchmark problems \cite{Mcdermott2012Genetic, senlikeana, Nicolau2015Guidelines} and 2 UCI data sets \cite{Dua:2017}. The Poly10 function is from \cite{poly10}.
All the benchmark problems are listed in Table \ref{tab:benchmarkfunctions}. Function Poly10 and Sphere5 use the function set below
\begin{equation}
\nonumber
    \{+,\ -,\ *,\ /\}.
\end{equation}
The other functions use the function set below
\begin{equation}
\nonumber
    \{+,\ -,\ *,\ /, \ \sin,\ \cos,\ e^n,\ \ln (\left| n \right|) \}.
\end{equation}
The division is protected by $f = x/(y+\varepsilon )$, where $\varepsilon$ is a very small number (e.g., 1E-100). Other benchmark details are listed in Table \ref{tab:benchmarkfunctions}.
All these difficult benchmark problems are commonly used due to their unique structural complexities with respect to objective formula. Several large scale benchmarks (e.g., 10 variables) for symbolic regression are considered one of the hard cases due to the difficulty of finding the solution in larger search space.

\begin{table*}[htb!]
\scriptsize
  \centerfloat
  \caption{Test problems used in this paper. $U[a,\ b,\ c]$ is $c$ samples uniformly randomized in $[a, \ b]$ for the variable. $E[a,\ b,\ c]$ are mesh points which are spaced equally with an interval of $c$, from $a$ to $b$ inclusive. }
  \label{tab:benchmarkfunctions}
  \begin{tabular}{lclll}
  \addlinespace
  \toprule
  Name & Variables & Function &  Training Set & Testing Set\\
  \midrule
  Sphere5 &5 &$x_1 ^2  + x_2 ^2  + x_3^2  + x_4^2  + x_5^2$& $U[1,\ 11,\ 1000]$& $U[1,\ 11,\ 1000]$ \\
  Dic1 &10 &$x_1  + x_2  + x_3  + x_4  + x_5  $& $U[1,\ 11,\ 1000]$& $U[1,\ 11,\ 1000]$ \\
  Dic3 &10 &$x_1  + \frac{{x_2 x_3 }}{{x_4 }} + \frac{{x_3 x_4 }}{{x_5 }}$& $U[1,\ 11,\ 1000]$& $U[1,\ 11,\ 1000]$  \\
  Dic4 &10 &$x_1 x_2  + x_2 x_3  + x_3 x_4 x_5  + x_5 x_6
  $& $U[1,\ 11,\ 1000]$& $U[1,\ 11,\ 1000]$  \\
  Dic5 &10 &$\sqrt {x_1 }  + \sin \left( {x_2 } \right) + \log _e \left( {x_3 } \right)$& $U[1,\ 11,\ 1000]$& $U[1,\ 11,\ 1000]$  \\
  Nico9 &2 &$x_1 ^4  - x_1^3  + x_2^2 /2 - x_2 $& $U[-5,\ 5,\ 1000]$& $U[-5,\ 5,\ 1000]$ \\
  Nico14 &6 &$\left( {x_5 x_6 } \right)/\left( {x_1 /x_2 x_3 /x_4 } \right)$& $U[-5,\ 5,\ 1000]$& $U[-5,\ 5,\ 1000]$ \\
  Nico16 &4 &$32 - 3\frac{{\tan \left( {x_1 } \right)}}{{\tan \left( {x_2 } \right)}}\frac{{\tan \left( {x_3 } \right)}}{{\tan \left( {x_4 } \right)}}$& $U[-5,\ 5,\ 1000]$& $U[-5,\ 5,\ 1000]$ \\
  Nico20 &10 &$\sum\limits_{i = 1}^5 {\frac{1}{{x_i }}}$& $U[-5,\ 5,\ 1000]$& $U[-5,\ 5,\ 1000]$ \\
  Poly10 &10 & \tabincell{l}{ $x_1 x_2  + x_2 x_3  + x_3 x_4  + x_4 x_5  + x_5 x_6$ \\ $+ x_1 x_7 x_9  + x_3 x_6 x_{10}$ } & $U[-1,\ 1,\ 250]$& $U[-1,\ 1,\ 250]$\\
  Pagie1 &2 &$\frac{1}{{1 + x_1^{ - 4} }} + \frac{1}{{1 + x_2^{ - 4} }}$& $E[-5,\ 5,\ 0.4]$ & $E[-4.95,\ 5.05,\ 0.4]$\\
  Nguyen6 &1 &$\sin \left( x \right) + \sin \left( {x + x^2 } \right) $& $U[-1,\ 1,\ 20]$& $U[-1,\ 1,\ 20]$\\
  Nguyen7 &1 &$\ln \left( {x + 1} \right) + \ln \left( {x^2  + 1} \right) $& $U[0,\ 2,\ 20]$& $U[0,\ 2,\ 20]$ \\
  Vlad3 &2 &$e^{ - x} x^3 (\cos x \sin x)(\cos x \sin ^2 x - 1)(y - 5) $ &  \tabincell{l}{$x:E[0.05,\ 10,\ 0.1]$ \\ $y:E[0.05,\ 10.05,\ 2]$}&  \tabincell{l}{$x:E[-0.5,\ 10.5,\ 0.05]$ \\ $y:E[-0.5,\ 10.5,\ 0.5]$} \\
  Energy &8 & Energy efficiency of buildings&  \\
  Concrete &8 & Concrete compressive strength&  \\
  \bottomrule

\end{tabular}

\end{table*}

\subsection{Compared Algorithm Configurations}

Standard GEP and GP \cite{geneticprog}, which both are classic and powerful models in the field of genetic programming methods, are compared with the three instances of the proposed method (marked as GA-NEEP, PSO-NEEP, and CMAES-NEEP). For a fair comparison, all common parameters in the listed methods are initialized with the same value. All the algorithms in the experiments used a population size of 100, and the number of generations 500. Other parameters of GA, PSO and CMA-ES were specified by default. For GP, we used tournament size of 3, maximum tree depth of 10, maximum tree length of 61, maximum mutation depth of 4, maximum crossover of depth 10, maximum grow depth of 1 and minimum grow depth of 1. For GEP, we used header length of 30, a crossover rate of 0.7, mutation rate of 0.1, IS transposition of 0.1, RIS transposition of 0.1, and the inversion rate of 0.1. For the three proposed methods (GA-NEEP, PSO-NEEP and CMAES-NEEP), we used header length of 30, hidden neurons 40, time steps of 10, the initial fixed weights sparsity of 0.5 and the initial optimizing weight range of [-2, 2].


\subsection{Results and Discussions}
\begin{table*}[hbt!]
\scriptsize
  \centerfloat
  \caption{Median, standard deviation and corresponding ranks of testing errors of the five compared algorithms. All differences are statistically significant according to a Wilcoxon test with a confidence level of $95\%$. Symbols $-$ and $+$ represent that the proposed method is respectively significantly worse than and better than the other two methods (GP and GEP). The other cases are marked with $=$. }
    \begin{tabular}{llllll}
    \addlinespace
    \toprule
    	   & GEP & GP  & GA-NEEP & PSO-NEEP & CMAES-NEEP \\
    \midrule
   Sphere5&	4.87e+04$\pm$6.58e+07&	\textbf{3.93e+02}$\pm$1.40e+02&	7.71e+02$\pm$4.46e+02&	6.27e+02$\pm$4.35e+02&	6.30e+02$\pm$1.17e+02	\\
rank& 	5&	1&	4&	2&	3	\\
& 	&	&	=&	=&	=	\\
 Dic1&	6.00e+02$\pm$4.91e+07&	1.55e+01$\pm$1.30e+01&	2.04e+01$\pm$1.58e+01&	2.83e+00$\pm$1.40e+01&	\textbf{4.97e-30}$\pm$7.67e-02	\\
rank& 	5&	3&	4&	2&	1	\\
& 	&	&	=&	+&	+	\\
 Dic3&	4.96e+02$\pm$5.58e+15&	1.29e+02$\pm$2.61e+01&	1.44e+02$\pm$2.39e+01&	\textbf{1.18e+02}$\pm$2.51e+01&	1.20e+02$\pm$1.30e+02	\\
rank& 	5&	3&	4&	1&	2	\\
& 	&	&	=&	+&	+	\\
 Dic4&	7.00e+04$\pm$1.28e+11&	6.68e+03$\pm$1.40e+04&	3.01e+04$\pm$1.50e+04&	5.12e+03$\pm$1.66e+04&	\textbf{3.62e+03}$\pm$1.79e+03	\\
rank& 	5&	3&	4&	2&	1	\\
& 	&	&	=&	=&	+	\\
 Dic5&	6.80e+00$\pm$1.35e+19&	8.96e-01$\pm$8.70e-01&	1.00e+00$\pm$2.61e-01&	5.99e-01$\pm$2.50e-01&	\textbf{5.54e-01}$\pm$1.21e-01	\\
rank& 	5&	3&	4&	2&	1	\\
& 	&	&	=&	+&	+	\\
 Nico9&	4.80e+04$\pm$1.85e+19&	\textbf{3.64e+02}$\pm$6.39e+03&	1.32e+04$\pm$2.07e+05&	1.27e+03$\pm$3.80e+03&	2.65e+03$\pm$5.62e+04	\\
rank& 	5&	1&	4&	2&	3	\\
& 	&	&	=&	=&	=	\\
 Nico14&	1.18e+07$\pm$1.47e+19&	1.20e+07$\pm$6.95e+07&	\textbf{1.18e+07}$\pm$1.12e+07&	1.18e+07$\pm$1.93e+10&	1.18e+07$\pm$5.89e+06	\\
rank& 	3&	5&	1&	4&	2	\\
& 	&	&	=&	=&	=	\\
 Nico16&	4.47e+09$\pm$1.20e+18&	4.48e+09$\pm$2.74e+11&	\textbf{4.46e+09}$\pm$3.69e+11&	4.91e+09$\pm$5.25e+12&	4.46e+09$\pm$1.36e+13	\\
rank& 	3&	4&	1&	5&	2	\\
& 	&	&	=&	-&	=	\\
 Nico20&	7.54e+02$\pm$1.21e+19&	1.86e+03$\pm$4.64e+04&	6.85e+02$\pm$2.15e+04&	\textbf{5.19e+02}$\pm$5.52e+04&	6.80e+02$\pm$1.54e+06	\\
rank& 	4&	5&	3&	1&	2	\\
& 	&	&	+&	+&	=	\\
 Poly10&	5.48e-01$\pm$7.29e+00&	3.24e-01$\pm$5.24e-02&	3.21e-01$\pm$6.11e-02&	3.21e-01$\pm$3.48e-02&	\textbf{3.17e-01}$\pm$2.83e-02	\\
rank& 	5&	4&	2&	3&	1	\\
& 	&	&	=&	+&	+	\\
 Pagie1&	9.61e-01$\pm$1.58e+19&	1.26e-01$\pm$1.14e-01&	1.95e-01$\pm$4.03e-02&	1.24e-01$\pm$3.42e-02&	\textbf{1.21e-01}$\pm$2.58e-02	\\
rank& 	5&	3&	4&	2&	1	\\
& 	&	&	=&	=&	=	\\
 Nguyen6&	2.10e-01$\pm$2.57e+19&	1.54e-01$\pm$1.61e-01&	1.10e-01$\pm$1.22e-01&	1.40e-02$\pm$3.28e-02&	\textbf{4.41e-03}$\pm$1.47e-02	\\
rank& 	5&	4&	3&	2&	1	\\
& 	&	&	+&	+&	+	\\
 Nguyen7&	2.63e-01$\pm$1.41e+19&	3.92e-02$\pm$1.35e-01&	3.30e-02$\pm$7.07e-01&	2.19e-03$\pm$8.33e-02&	\textbf{1.15e-03}$\pm$4.86e-03	\\
rank& 	5&	4&	3&	2&	1	\\
& 	&	&	+&	+&	+	\\
 Vlad3&	7.63e+00$\pm$2.57e+20&	1.22e+00$\pm$1.52e+12&	\textbf{9.47e-01}$\pm$Inf&	1.05e+00$\pm$Inf&	1.01e+00$\pm$1.04e+33	\\
rank& 	5&	4&	1&	3&	2	\\
& 	&	&	+&	=&	+	\\
 Energy&	1.06e+02$\pm$3.81e+18&	2.58e+01$\pm$3.78e+01&	4.52e+01$\pm$2.20e+01&	\textbf{2.17e+01}$\pm$7.48e+00&	2.34e+01$\pm$6.57e+00	\\
rank& 	5&	3&	4&	1&	2	\\
& 	&	&	=&	+&	+	\\
 Concrete&	3.39e+02$\pm$5.78e+18&	2.26e+02$\pm$6.63e+01&	2.21e+02$\pm$2.66e+01&	1.80e+02$\pm$3.41e+01&	\textbf{1.66e+02}$\pm$3.38e+01	\\
rank& 	5&	4&	3&	2&	1	\\
& 	&	&	=&	+&	+	\\
Avg. Rank & 4.69 & 3.38 & 3.06 & 2.25 & 1.63 \\
\bottomrule

    \end{tabular}
  \label{tab:results}

\end{table*}
\par
Table \ref{tab:results} summarizes the test errors obtained by GEP, GP and the three versions of NEEP on all the benchmark problems. The median and standard deviation are summarized over the 50 independent repeated trials for each of the 16 benchmarks function. It can be observed that the proposed methods (GA-NEEP, PSO-NEEP, CMAES-NEEP) significantly outperformed GEP and GP on 14 out of 16 problems according to the median of MSE, and perform competitively on the remaining problems. In particular, CMAES-NEEP reported dramatically lower MSE and more stable performance (according to their standard deviation values) on all the high dimensional data (Poly10, Dic1, Dic3, Dic4, Dic5, Nico20), while GEP and GP failed to locate the global optimum for these problems because the solution expressions of a high dimensional problem become overwhelming or extremely complicated. Therefore, such problems may become tough for the traditional GEP and GP due to their lack of capability to encode a complex function in a single string.

\par
\begin{figure*}[htb]
    \centering
    \includegraphics[width=0.97\textwidth]{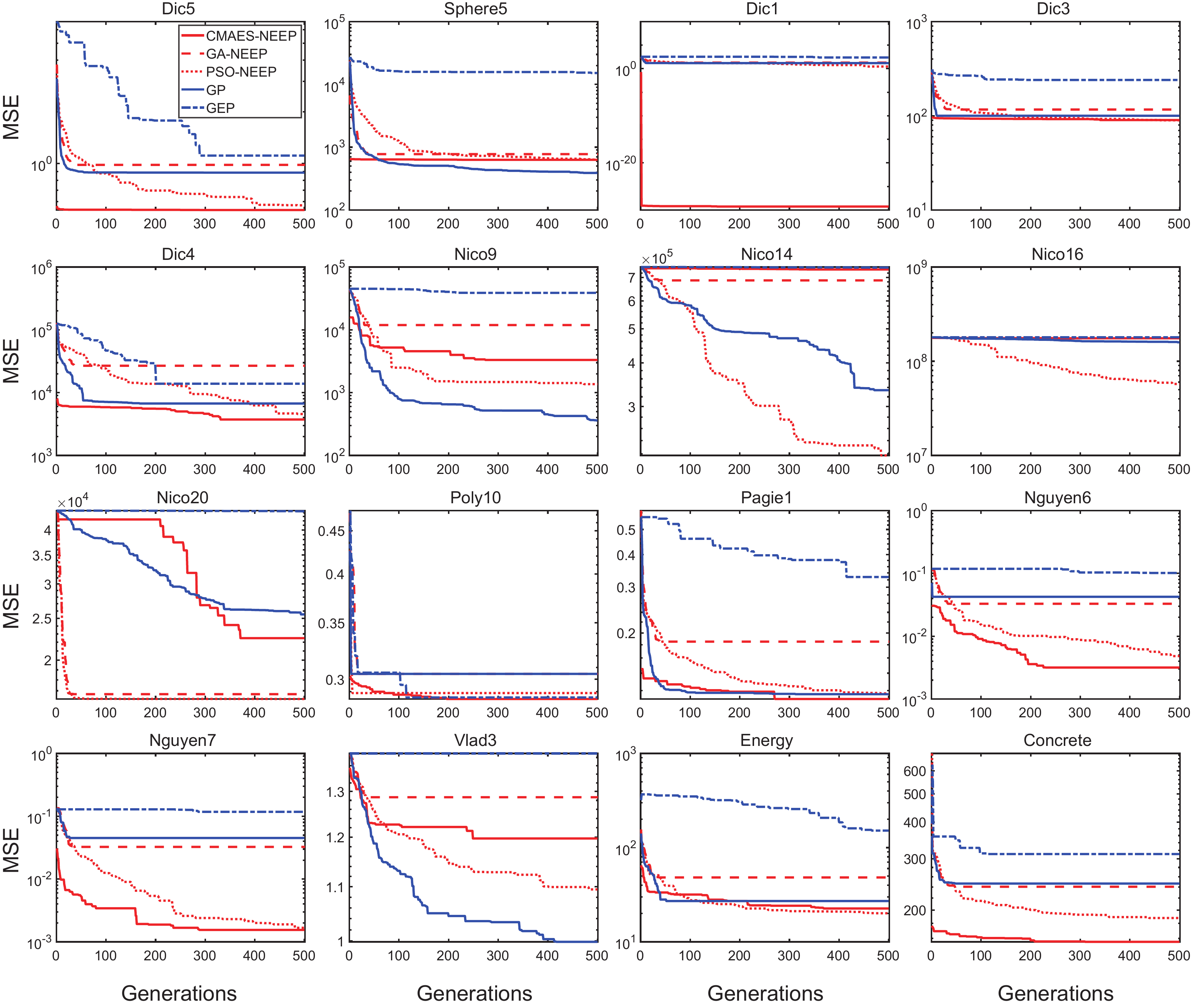}
    \caption{Evolution of the average best training errors of 50 independent trials for all compared algorithms. }
    \label{fig:convergence}
\end{figure*}

For the two regression data sets Concrete and Energy, the convergence curves in Fig. \ref{fig:convergence} and test errors in Table \ref{tab:results} reveal that CMAES-NEEP and PSO-NEEP have remarkable performance and high stability among all the compared methods. According to the convergence curves, in some functions (Nico16, Dic1, Dic3) these methods illustrate premature convergence and get stuck at a local optimum during evolution. For Nico9 and Sphere5 problem, GP sits as the best method among all the compared algorithms, due to its property of reusability of existing nodes during the encoding of the expression tree. On the other hand, GEP stands at the worst, and all other proposed methods show the competitive results concerning GP.

\par
According to the test error results and the convergence curves, we can observe that the proposed method has shown more stable performance and faster training speed on most of the listed benchmark functions. In addition, we can obtain a rough conclusion that among the three evolutionary algorithms, CMA-ES is the most powerful, PSO is the second, and GA is the last one. These ranks also conform the general impression of their performances on many artificial benchmark functions in evolutionary computation. Therefore, it is important for NEEP to choose a strong optimizer for searching better neural networks.

\section{Conclusion}

This study proposes a novel continuous neural encoding approach to improve conventional linear representation in genetic programming methods for solving symbolic regression. Linear representation methods manipulate the expression tree structures in a discrete manner, which does not assist in a localized search of solution space.
The neuro-encoded expression programming (NEEP) transforms the combinatorial problem to a continuous problem by using a neural network to generate an expression string, thus powerful numerical optimization method can be adopted to find a better mathematical function for symbolic regression. Empirical analysis demonstrates the method has the potential to deliver improved test accuracy and efficiency.
\par
There are several interesting future research directions, such as to explore more neural network architectures for encoding and introduce the constant creation in string encoding mechanism.
This new framework for now only improves one of linear representation methods and focuses on one application in spite of its potential for applying on more methods and applications. Therefore, one of the future works is to explore other types of genetic programming methods with neural networks. Another consideration is to apply NEEP to more applications (e.g., classification, digital circuit design and path planning).

\section*{Acknowledgments}
This work was supported by National Natural Science Foundation of China under Grant No. 61573166, No. 61572230, No. 61872419, No. 61873324, No. 81671785, No. 61672262. Project of Shandong Province Higher Educational Science and Technology Program under Grant No. J16LN07. Shandong Provincial Natural Science Foundation No. ZR2019MF040, No. ZR2018LF005. Shandong Provincial Key R\&D Program under Grant No. 2018GGX101048, No. 2016ZDJS01A12, No. 2016GGX101001, No. 2017CXZC1206. Taishan Scholar Project of Shandong Province, China.

%
%
%
\bibliographystyle{splncs04}
\bibliography{ref}
%
%
%
%
%
\end{document}